\documentclass{article}
\usepackage{spconf,amsmath,graphicx}

\usepackage{amssymb}
\usepackage{subfig}
\usepackage{pbox}
\DeclareMathOperator*{\argmin}{arg\,min}
\DeclareMathOperator*{\argmax}{arg\,max}

\title{Class-Specific Poisson Denoising by Patch-Based Importance Sampling}
\name{Milad Niknejad, Jos\'e M. Bioucas-Dias, M\'ario A. T. Figueiredo\thanks{The research leading to these results has been funded by the European Union's  7-th Framework Programme (FP7-PEOPLE-2013-ITN), grant agreement 607290 (SpaRTaN), and from {\it Funda\c{c}\~ao para a Ci\^encia e Tecnologia} (FCT), grants UID/EEA/5008/2013 and PTDC/EEI-PRO/0426/2014.}}
\address{Instituto de Telecomunicacoes\\
Instituto Superior Tecnico\\
Universidade de Lisboa, Portugal\\}
\begin{document}
%
\maketitle
\begin{abstract}
In this paper, we address the problem of recovering images degraded by Poisson noise, where the image is known to belong to a specific class. In the proposed method, a dataset of clean patches from images of the class of interest is clustered using multivariate Gaussian distributions. In order to recover the noisy image, each noisy patch is assigned to one of these distributions, and the corresponding minimum mean squared error (MMSE) estimate is obtained. We propose to use a self-normalized importance sampling approach, which is a method of the Monte-Carlo family, for the both determining the most likely distribution and approximating the MMSE estimate of the clean patch. Experimental results shows that our proposed method outperforms other methods for Poisson denoising at a low SNR regime.

\end{abstract}
\section{Introduction}
\label{sec:intro}
Recovering noisy images is a fundamental problem in image processing and computer vision. In many image denoising applications, especially in low intensity (photon-limited) images, the noise follows a Poisson distribution. Photon-limited acquisition scenarios are often found in astronomical and medical imaging, and in many other areas. 

The classical formulation for Poisson denoising considers the task of recovering an underlying image, the pixels of which are stacked in a vector $\mathbf{x} \in \mathbb{R}_+^{N}$, from Poissonian observations $\mathbf{y} \in \mathbb{N}_0^{N}$, \textit{i.e.} 
\begin{equation}
\mathbb{P}(\mathbf{y}|\mathbf{x}) = \prod_{j=1}^N \frac{e^{-\mathbf{x}_{[j]}}\mathbf{x}_{[j]}^{\mathbf{y}_{[j]}}}{\mathbf{y}_{[j]}!}.\label{eq:poisson}
\end{equation}

A common approach to Poisson densoising proceeds as follows: first, a \textit{variance stabilization transform} (VST) is applied (\textit{e.g.}, the Anscombe transform \cite{1948_anscombe_transformation}), which approximately turns the Poisson distribution into a Gaussian of unit variance. Then, one of the many existing methods for Gaussian denoising is applied to the transformed image. Finally an inverse transformation yields the image estimate \cite{2009_dupe_proximal,2011_makitalo_optimal}. However, this approach is inaccurate for low  intensity (low SNR) images, thus many methods denoise the Poissonian images without any transformation \cite{2014_salmon_poisson}. Our method belongs to this second category.

In many applications, the image to be recovered is known to belong to a certain class, such as text, face, or fingerprints. Knowing this information can help to learn a statistical prior that is better adapted to that specific class than a generic prior, and this prior can be learned from a dataset of clean images from that class. Recently this class-specific approach has been addressed in \cite{2015_luo_adaptive,2016_teodoro_image}, for denoising and deblurring under Gaussian noise.

Some patch-based methods for recovering images degraded by Gaussian noises use multivariate priors for image restoration to obtain \textit{maximum  a posteriori} (MAP) or \textit{minimum mean squared error} (MMSE) estimates of the clean image. Most of these methods fit a mixture multivariate Gaussian distributions to patches, either from an  external dataset \cite{2011_zoran_learning}, or from the noisy image itself \cite{Teodoro2015,Niknejad_TIP2015}. Although for Poisson denoising, patch-based methods have been proposed \cite{2014_salmon_poisson,Giryes2014},  there is a lack of methods that use an explicit prior on image patches, due to the non-Gaussian forward model, which makes it difficult to compute the corresponding MAP or MMSE estimates. On the other hand, the fitted Gaussian mixture prior is an approximation of the true distribution that characterizes the image patches. 

In this paper, we propose a class-adapted method to recover images degraded by Poisson noise using a dataset of clean images of the same class.  In our method, a mixture of multivariate Gaussian distribution is fitted to the external dataset of clean image patches. For each noisy patch of the observed image, we use the self-normalized importance sampling \cite{Hesterberg}, which is an approach from the family of Monte-Carlo methods, to both choose the proper prior distribution and approximate the MMSE estimation of each patch. Using importance sampling allows us to approximate the MMSE estimate using the true distribution of image patches, if the samples are the patches of clean images in the dataset. Furthermore, our proposed method can be extended to other forward models with known noise distribution, and different prior densities of clean patches.

In the following sections, we first briefly review the self-normalizing importance sampling approach. Then, we describe the proposed method for MMSE Poisson image denoising based on self-normalizing importance sampling. Section \ref{sec:expre} reports experimental results.

\section{Self-Normalized Importance sampling}
Our approach uses a technique from the family of Monte-Carlo methods, called \textit{self-normalized importance sampling} (SNIS) \cite{Hesterberg,mcbook}, which we will now briefly review. Assume that the objective is to compute (or approximate) $\mathbb{E}[f({\bf X})]$, the expectation of a function $f:\mathbb{R}^d \to \mathbb{R}^p$ of a random variable $X\in\mathbb{R}^d$. Denoting the support of the random variable $\mathbf{X}$ as $\mathcal{R} \subset \mathbb{R}^d$, the expectation is given by
\begin{equation}
\mathbb{E}[f(\mathbf{x})]=\int_{\mathcal{R}} f(\mathbf{x}) p(\mathbf{x}) d\mathbf{x}.
\end{equation}

Consider also that only an un-normalized version  $\tilde{p}(\mathbf{x}) = c \, p({\bf x})$ of the probability density function $p(\mathbf{x})$ of ${\bf X}$ is known, where $c$ is unknown. Let $\tilde{q}(\mathbf{x}) = d\, q({\bf x})$ be another un-normalized density, where the normalizing constant $d$ is also unknown, but from which samples can be more efficiently obtained than from  $p({\bf x})$.

SNIS produces an estimate of $\mathbb{E}[f({\bf X})]$ given by
\begin{equation}
\hat{\mathbb{E}}_n[f(\mathbf{x})]=\frac{\displaystyle \sum_{j=1}^{n}{f(\mathbf{x}_j) w(\mathbf{x}_j)}}{\displaystyle \sum_{j=1}^{n}{w(\mathbf{x}_j)}},\label{eq:slfimps}
\end{equation}
where $w(\mathbf{x}_j)=\frac{\tilde{p}(\mathbf{x}_j)}{\tilde{q}(\mathbf{x}_j)}$, and $\mathbf{x}_1,...,\mathbf{x}_n$ are $n$ independent and identically distributed samples from distribution $q(\mathbf{x})$ \cite{Hesterberg,mcbook}. This approximation can be shown to converges to the true value of $\mathbb{E}[f(\mathbf{x})]$ as $n$ goes to infinity.

\section{Proposed Method}
\label{sec:format}
\subsection{Prior Learning}
The first step of the proposed method is to fit a set of $K$ multivariate Gaussians to the patches in the external dataset of clean images. We adopt the 
so-called \textit{classification EM} (CEM) algorithm \cite{Celeux}, rather than a standard EM algorithm as in \cite{Teodoro2015,2016_teodoro_image,2011_zoran_learning}, for reasons explained below. The CEM algorithm works by alternating between the following two steps (after being initialized by standard $K$-means clustering):
\begin{enumerate}
\item From the set of patches assigned to each cluster $k \in \{1,...,K\}$, denoted $\mathbf{X}_k$, obtain estimates of the mean $\mu_k$ and covariance matrix $\Sigma_k$ of the corresponding Gaussian density, which are simply the sample mean and the sample covariance of the patches in $\mathbf{X}_k$.
\item Assign each patch to the cluster under which it has the highest likelihood, that is, patch $\mathbf{x}_j$ is assigned to $\mathbf{X}_k$ if
\[
k = \arg\max_{m} \mathcal{N}(\mathbf{x}_j;\mu_m,\Sigma_m),
\]
where, as usual,  $\mathcal{N}(\mathbf{x};\mu,\Sigma)$ denotes a Gaussian density of mean $\mu$ and covariance $\Sigma$, computed at $\mathbf{x}$. 
\end{enumerate}
In the denoising step, each noisy patch will be assigned to one of these clusters. The reason for this kind of clustering is that in the simple importance sampling approach, for a fixed number of samples $n$, the MSE of the estimator is proportional to the variance of samples being averaged \cite{mcbook}. Similarly in the multivariate case, for a fixed $n$, it has been shown that the MSE of the estimator of a particular entry in the vector decreases as the variance of the samples of that entry, given the noisy patch, decreases \cite{2012_levin_patch}. Consequently, since in practice we use a limited number of patches from the external dataset, by clustering this way we expect to reduce the estimator variance, without increasing too much the number of samples.

It should be noted that the above procedure needs to be applied once for a given dataset of class-specific images. 


\subsection{Image denoising}
In the denoising step, each patch is assigned to one of the clusters obtained in the learning stage. However, we  only have noisy patches, thus the assignment is not trivial. If the noise was Gaussian, the assignments could be made in closed-form, but this is not the case with Poisson observations. 
Our main contribution is a new method, based on SNIS, to simultaneously determine the cluster and estimate the clean patch. 

Define the random variable $k_i \in \{1,...,K\}$, which indicates the cluster to which the $i$-th patch belongs, and denote the corresponding distribution as $p(\mathbf{x}_i|k_i)$. Having a set of learned cluster distributions, our objective is to solve the following simultaneous classification and MMSE estimation problem, given a noisy patch $\mathbf{y}_i$:
\begin{equation}
\label{eq:main}
(\hat{\mathbf{x}}_i,\hat{k}_i) = \argmin_{(\mathbf{u},k)}{\int_{\mathbb{R}_+^m}\| \mathbf{u} -\mathbf{x}\|_2^2 \; p(\mathbf{x}|\mathbf{y}_i,k)\, \; d\mathbf{x}} ,
\end{equation}
where $m$ is the number of pixels in each patch. In other words, we seek the estimate and the cluster that yield the minimum MSE. As shown below, we solve the above problem by the alternating minimization approach, but first we need to address the problem of how to approximate this integral.

First,  using Bayes rule and the fact that $p(\mathbf{y}_i|\mathbf{x},k) = p(\mathbf{y}_i|\mathbf{x})$ (\textit{i.e.}, given the clean patch, the noisy one does not depend on the cluster), the integral in (\ref{eq:main}) can be written as
\begin{equation}
\mathbb{E}[\| {\bf x} - {\bf u}\|_2^2 | {\bf y}_i, k] = \int_{\mathbb{R}_+^m}\| \mathbf{u}-\mathbf{x} \|_2^2 \; \frac{p(\mathbf{y}_i|\mathbf{x})\, p(\mathbf{x}|k)}{p(\mathbf{y}_i|k)} \; d\mathbf{x}.
\end{equation}
Then, using SNIS, the above integral can be approximated by
\begin{equation}
\label{eq:intapp}
\hat{\mathbb{E}}_{n}[\| {\bf x} - {\bf u}\|_2^2 | {\bf y}_i, k] = \frac{\displaystyle \sum_{j=1}^{n} \|\mathbf{u}-\mathbf{x}_j\|_2^2 \; p(\mathbf{y}_i|\mathbf{x}_j)}{\displaystyle \sum_{j=1}^{n} p(\mathbf{y}_i|\mathbf{x}_j)} 
\end{equation}
where the $\mathbf{x}_j$, for $j=1,...,n$ are samples from the distribution $p(\mathbf{x}|k)$. Exploiting the self-normalized importance sampling formula (\ref{eq:slfimps}), in (\ref{eq:intapp}) we considered $\tilde{p}(\mathbf{x})= p(\mathbf{y}_i|\mathbf{x})\, p(\mathbf{x}|k)$, with the unknown constant $c=1/p(\mathbf{y}_i|k)$ and $q(\mathbf{x})=p(\mathbf{x}|k)$. 

The minimization with respect  to $k$ in (\ref{eq:main}), while ${\bf u}$ is fixed to be $\hat{\mathbf{x}}_i$, can then be approximated as
\begin{equation}
\label{eq:cmpcluster}
\hat{k}_i = \argmin_{k} \hat{\mathbb{E}}_{n_2}[\| {\bf x} - {\hat{\mathbf{x}}_i}\|_2^2 | {\bf y}_i, k].
\end{equation}
Note that for computing \eqref{eq:cmpcluster}, $n_2$ samples from each distribution  $p(\mathbf{x}| k)$ are randomly extracted.

The next step towards an alternating minimization method is a way to minimize ${\mathbb{E}}_{n}[\| {\bf x} - {\bf u}\|_2^2 | {\bf y}_i, k]$ with respect to ${\bf u}$, for a given $k=\hat{k}_i$. This is the well-known MMSE estimation criterion, which is given by the posterior expectation $\mathbb{E}(\mathbf{x}|\mathbf{y}_i,\hat{k}_i)$. Computing this expectation cannot be done in closed-form, under the Poisson observation model and Gaussian prior $p({\bf x}|k) = \mathcal{N}({\bf x};\mu_k,\Sigma_k)$. To tackle this difficulty,  we resort again to SNIS, 
\begin{equation}
\label{eq:imd}
\hat{\mathbf{x}}_i = \hat{\mathbb{E}}_{n_1}[\mathbf{x}|\mathbf{y}_i,\hat{k}_i] = \frac{\displaystyle\sum_{j=1}^{n_1}{\mathbf{x}_j p(\mathbf{y}_i|\mathbf{x}_j)}}{\displaystyle \sum_{j=1}^{n_1}{p(\mathbf{y}_i|\mathbf{x}_j)}}
\end{equation}
where $\mathbf{x}_1,...,\mathbf{x}_{n_1}$ are samples drawn from the distribution  $p(\mathbf{x}| \hat{k})$. The approximation in (\ref{eq:imd}) has been used before for patch-based denoising \cite{2011_levin_natural}, but without noticing its connection to SNIS.

Since the multivariate Gaussian distributions fitted to the set of patches is only a crude approximation of the true distribution of the patches, samples from $p(\mathbf{x}|k)$ may yield poor results in the SNIS approximation. Instead of sampling from $p(\mathbf{x}|k)$, we directly sample form the set patches in the dataset assigned to that $k$-th cluster, $\mathbf{X}_k$, which turns out to yield better results.

We initialize our algorithm by $\mathbf{u}=\mathbf{y}_i$. Our algorithm then alternates between (\ref{eq:imd}) and (\ref{eq:cmpcluster}), to assign each patch to a cluster and obtain an MMSE estimate of the patch. The procedure of iterative clustering and denoising has been used in some well-known patch-based denoising methods \cite{2007_dabov_bm3d,2011_zoran_learning}. The patches are returned to the original position in the image and are averaged in the overlapping pixels.

Note that our method can be applied to any noise model with a known distribution. In this paper, we only consider Poissonian noise and we leave other possible noise distributions for future work. For the Poisson case, the conditional distribution of a noisy patch ${\bf y}$, given a clean patch $\mathbf{x} \in \mathbb{R}_+^m$, is as shown in  \eqref{eq:poisson}, with $N=m$.

\begin{figure}[!h]
\centering
\begin{tabular}{|p{8.0cm}|}
\hline
\begin{itemize}
\item Initialization: cluster the training patches into $K$ clusters, $\mathbf{X}^{(1)}_1 \ldots \mathbf{X}^{(1)}_K$, via the $k$-means algorithm
\item Main loop: for $i=1, \ldots ,I$
\begin{itemize}
\item Estimate the parameter of Gaussian distributions $\Theta_k^{(i)}=\{\boldsymbol{\mu}^{(i)}_k,\mathbf{\Sigma}^{(i)}_k\}$ as the sample mean and the sample covariance matrix of $\mathbf{X}^{(i)}_k$
\item For each patch ${\bf x}_j$, assign it to a cluster, {\it i.e.}, put ${\bf x}_j$ into $\mathbf{X}^{(i)}_{\hat{k}_j}$, where
\[
\hat{k}_j = \argmax_k \mathcal{N}( \mathbf{x}_j; \boldsymbol{\mu}_k^{(i)},\boldsymbol{\Sigma}_k^{(i)})
\]
\end{itemize}
\item Output: \{$\mathbf{X}_1^{(I)} \ldots \mathbf{X}_K^{(I)}$\}
\end{itemize}
\\
\hline
\end{tabular}
\caption{Learning the prior from a class-specific dataset.}
\label{fig:dbcluster}
\end{figure}


\section{Practical considerations and experimental results}
\label{sec:expre}
The reported results in this section were obtained on the Gore dataset of face images \cite{2012_peng_rasl} and on the dataset of text images used in \cite{2015_luo_adaptive}. For each dataset, $5$ images are randomly chosen as test images and the rest are chosen for training. For the face images, we extracted $95\times10^3$ patches from the training data, whereas from the text dataset, $75\times10^3$ patches were extracted.

The described method is computationally expensive, if it is applied to all patches in the dataset. However, the key to reduce the computational complexity is to limit the number of patch samples used for estimating the clean patches and determining the cluster i.e. $n_1$ and $n_2$. Our results in this section show that for a very limited number of samples, we obtain acceptable results, which outperforms other Poisson denoising methods for the tested datasets. For determining the cluster, we set $n_2=30$ which is overall $600$ patches for all $k=20$ clusters and is less than $1\%$ of the samples in each external datasets. The number of samples, $n_1$, used for denoising each patch  was set to $300$. So, overall for each noisy patch we processed $900$ patches which is in computational complexity roughly similar to an internal non-local denoising with the patches constrained in $30\times30$ window.
Unlike the original non-local means, which only the central pixel of each patch is denoised, in our method the whole patch is denoised by (\ref{eq:imd}), the patches are then returned to the original position in the image and are averaged in the overlapping pixels. In order to further reduce the computational complexity, we extract the patches from the noisy image every $2$ pixels along the row and the column of the image. 

In Table \ref{tb:face}, the average result of PSNR of the $5$ tested images for the face dataset are compared for different methods. The results of our method result from two iterations of the alternating minimization approach. We found that increasing the number of iteration to more than two does not noticeably increase the quality of obtained image estimate, while it obviously increases the computational cost. The reason may lie in the fact that the discrete variables $k_i$ computed in \eqref{eq:cmpcluster} stabilize after a couple of iterations.

\begin{table}[]
\centering
\caption{Denoising PSNR results (in dB) for different peak values of face images in the Gore face database \cite{2012_peng_rasl}. The results are averaged over 5 test images.}
\label{tb:face}
\begin{tabular}{|l|l|l|l|l|l|l|}
\hline
  & $2$ & $5$ & $10$ & $15$\\ \hline
NL-PCA \cite{2014_salmon_poisson}& 19.69 & 22.87&23.80&25.01  \\ \hline
VST+BM3D \cite{2011_makitalo_optimal}&20.80&23.15&	24.79&25.41 \\ \hline
Poisson NL-means \cite{2010_deledalle_poisson}& 21.12&23.41 &24.73&25.32  \\ \hline
P4IP \cite{2016_rond_poisson}&20.03&23.78&24.88&25.84 \\ \hline
Our method & \bf{21.31}&\bf{23.95}&\bf{25.78}&\bf{27.40}  \\ \hline
\end{tabular}
\end{table}

\begin{figure}[!t]
\subfloat[]{\includegraphics{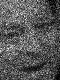}%
\label{(a)}}~
\subfloat[]{\includegraphics{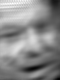}%
\label{b)}}~
\subfloat[]{\includegraphics{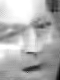}%
\label{(c)}}~
\\
\hfill
\subfloat[]{\includegraphics{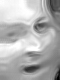}%
\label{(d)}}~
\subfloat[]{\includegraphics{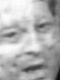}%
\label{(e)}}~
\subfloat[]{\includegraphics{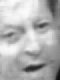}%
\label{(f)}}
\hfill
\caption{An example of denoising of a face image using the Gore dataset ; (a) Noisy image (Peak value=$10$); (b) Non-local PCA (PSNR=22.60); (c) VST+BM3D (PSNR=24.79); (d) Poisson non-local means (PSNR=24.55); (e) Proposed (First iteration) (PSNR=25.88); (f) Proposed (Second iteration) (PSNR=26.40).}
\label{fig:face}
\end{figure}

\begin{figure}[!h]
\subfloat[]{\includegraphics[width=.16\textwidth]{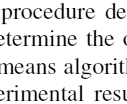}%
\label{(a)}}~
\subfloat[]{\includegraphics[width=.16\textwidth]{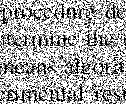}%
\label{(b)}}~
\subfloat[]{\includegraphics[width=.16\textwidth]{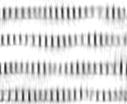}%
\label{c)}}~
\\
\hfill
\subfloat[]{\includegraphics[width=.16\textwidth]{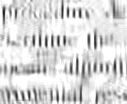}%
\label{(d)}}~
\subfloat[]{\includegraphics[width=.16\textwidth]{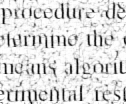}%
\label{(e)}}~
\subfloat[]{\includegraphics[width=.16\textwidth]{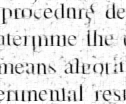}%
\label{(f)}}
\hfill
\caption{Comparison of denoising of general methods with our class-specific one for recovering the text images; (a) Original image (b) Noisy image (Peak value=$2$) (c) Non-local PCA (PSNR=14.95) (d) VST+BM3D (PSNR=14.55); (e) Proposed (First iteration) (PSNR=17.21); (f) Proposed (Second iteration) (PSNR=18.64).}
\label{fig:text}
\end{figure}

In Fig. (\ref{fig:face}), an example of denoising results for the face images with the peak value of $10$ using the dataset in \cite{2012_peng_rasl} is illustrated. It can be seen that  our method also improves noticeably the visuals quality of image.

An example of denoising text images is shown in Fig. (\ref{fig:text}). It can be seen that the general image denoising methods fail to reconstruct the true image properly. However our class-specific method outperforms by a relatively high margin. 

\section{Conclusion}
We proposed a method for class-specific denoising of images degraded by Poisson noise. We proposed to use a method from the Monte-Carlo family,  called self-normalized importance sampling, in order to determine the prior and approximate the MMSE estimation. The sampling approach allowed us to approximate the MMSE using the true underlying priors rather than fitted multivariate Gaussian distributions. Our results showed that our method outperforms other methods in both PSNR and visual quality.

------------------------------------
\bibliographystyle{IEEEbib}
\bibliography{strings,refs}

\begin{thebibliography}{10}

\bibitem{1948_anscombe_transformation}
F.~Anscombe,
\newblock ``The transformation of {Poisson}, binomial and negative-binomial
  data,''
\newblock {\em Biometrika}, vol. 35, no. 3/4, pp. 246--254, 1948.

\bibitem{2009_dupe_proximal}
F.~Dup{\'e}, J.~Fadili, and J.~Starck,
\newblock ``A proximal iteration for deconvolving poisson noisy images using
  sparse representations,''
\newblock {\em IEEE Transactions on Image Processing}, vol. 18, no. 2, pp.
  310--321, 2009.

\bibitem{2011_makitalo_optimal}
M.~Makitalo and A.~Foi,
\newblock ``Optimal inversion of the {Anscombe} transformation in low-count
  {Poisson} image denoising,''
\newblock {\em IEEE Transactions on Image Processing}, vol. 20, no. 1, pp.
  99--109, 2011.

\bibitem{2014_salmon_poisson}
J.~Salmon, Z.~Harmany, C.~Deledalle, and R.~Willett,
\newblock ``Poisson noise reduction with non-local {PCA},''
\newblock {\em Journal of Mathematical Imaging and Vision}, vol. 48, no. 2, pp.
  279--294, 2014.

\bibitem{2015_luo_adaptive}
E.~Luo, S.~Chan, and T.~Nguyen,
\newblock ``Adaptive image denoising by targeted databases,''
\newblock {\em IEEE Transactions on Image Processing}, vol. 24, no. 7, pp.
  2167--2181, 2015.

\bibitem{2016_teodoro_image}
A.~Teodoro, J.~Bioucas-Dias, and M.~Figueiredo,
\newblock ``Image restoration and reconstruction using variable splitting and
  class-adapted image priors,''
\newblock in {\em IEEE International Conference on Image Processing}, 2016.

\bibitem{2011_zoran_learning}
D.~Zoran and Y.~Weiss,
\newblock ``From learning models of natural image patches to whole image
  restoration,''
\newblock in {\em 2011 International Conference on Computer Vision}. IEEE,
  2011, pp. 479--486.

\bibitem{Teodoro2015}
A.~Teodoro, M.~Almeida, and M.~Figueiredo,
\newblock ``Single-frame image denoising and inpainting using {Gaussian}
  mixtures,''
\newblock in {\em 4th International Conference on Pattern Recognition
  Applications and Methods}, 2015.

\bibitem{Niknejad_TIP2015}
M.~Niknejad, H.~Rabbani, and M.~Babaie-Zadeh,
\newblock ``Image restoration using {Gaussian} mixture models with spatially
  constrained patch clustering,''
\newblock {\em IEEE Transactions on Image Processing}, vol. 24, pp. 3624--3636,
  2015.

\bibitem{Giryes2014}
R.~Giryes and M.~Elad,
\newblock ``Sparsity-based {Poisson} denoising with dictionary learning,''
\newblock {\em IEEE Transactions on Image Processing}, vol. 23, no. 12, pp.
  5057--5069, 2014.

\bibitem{Hesterberg}
T.~Hesterberg,
\newblock ``Weighted average importance sampling and defensive mixture
  distributions,''
\newblock {\em Technometrics}, vol. 37, no. 2, pp. 185--192, 1995.

\bibitem{mcbook}
A.~Owen,
\newblock {\em Monte Carlo Theory, Methods and Examples},
\newblock 2013,
\newblock Available at {http://statweb.stanford.edu/~owen/mc/}.

\bibitem{Celeux}
G.~Celeux and G.~Govaert,
\newblock ``A classification {EM} algorithm for clustering and two stochastic
  versions,''
\newblock {\em Computational Statistics and Data Analysis}, vol. 14, no. 3, pp.
  315--332, 1992.

\bibitem{2012_levin_patch}
Anat Levin, Boaz Nadler, Fredo Durand, and William~T Freeman,
\newblock ``Patch complexity, finite pixel correlations and optimal
  denoising,''
\newblock in {\em European Conference on Computer Vision}. Springer, 2012, pp.
  73--86.

\bibitem{2011_levin_natural}
A.~Levin and B.~Nadler,
\newblock ``Natural image denoising: Optimality and inherent bounds,''
\newblock in {\em Computer Vision and Pattern Recognition (CVPR), 2011 IEEE
  Conference on}. IEEE, 2011, pp. 2833--2840.

\bibitem{2007_dabov_bm3d}
Kostadin Dabov, Alessandro Foi, Vladimir Katkovnik, and Karen Egiazarian,
\newblock ``Image denoising by sparse 3-d transform-domain collaborative
  filtering,''
\newblock {\em IEEE Transactions on image processing}, vol. 16, no. 8, pp.
  2080--2095, 2007.

\bibitem{2012_peng_rasl}
Y.~Peng, A.~Ganesh, J.~Wright, W.~Xu, and Y.~Ma,
\newblock ``Rasl: Robust alignment by sparse and low-rank decomposition for
  linearly correlated images,''
\newblock {\em IEEE Transactions on Pattern Analysis and Machine Intelligence},
  vol. 34, no. 11, pp. 2233--2246, 2012.

\bibitem{2010_deledalle_poisson}
C.~Deledalle, F.~Tupin, and L.~Denis,
\newblock ``Poisson nl means: Unsupervised non local means for poisson noise,''
\newblock in {\em Image processing (ICIP), 2010 17th IEEE int. conf. on}. IEEE,
  2010, pp. 801--804.

\bibitem{2016_rond_poisson}
A.~Rond, R.~Giryes, and M.~Elad,
\newblock ``Poisson inverse problems by the plug-and-play scheme,''
\newblock {\em Journal of Visual Communication and Image Representation}, vol.
  41, pp. 96--108, 2016.

\end{thebibliography}

\end{document}